# A Person Re-identification Data Augmentation Method with Adversarial Defense Effect

Yunpeng Gong, Zhiyong Zeng, Liwen Chen, Yifan Luo, Bin Weng, Feng Ye

College of Mathematics and Information, Fujian Normal University
{fmonkey625}@gmail.com

***Abstract.*** The security of the Person Re-identification(ReID) model plays a decisive role in the application of ReID. However, deep neural networks have been shown to be vulnerable, and adding undetectable adversarial perturbations to clean images can trick deep neural networks that perform well in clean images. We propose a ReID multi-modal data augmentation method with adversarial defense effect: 1) Grayscale Patch Replacement, it consists of Local Grayscale Patch Replacement(LGPR) and Global Grayscale Patch Replacement(GGPR). This method can not only improve the accuracy of the model, but also help the model defend against adversarial examples; 2) Multi-Modal Defense, it integrates three homogeneous modal images of visible, grayscale and sketch, and further strengthens the defense ability of the model. These methods fuse different modalities of homogeneous images to enrich the input sample variety, the variaty of samples will reduce the over-fitting of the ReID model to color variations and make the adversarial space of the dataset that the attack method can find difficult to align, thus the accuracy of model is improved, and the attack effect is greatly reduced. The more modal homogeneous images are fused, the stronger the defense capabilities is . The proposed method performs well on multiple datasets, and successfully defends the attack of MS-SSIM proposed by CVPR2020 against ReID [10], and increases the accuracy by 467 times(0.2% to 93.3%).The code is available at https://github.com/finger-monkey/ReID_Adversarial_Defense

## 1 Introduction

Although deep learning has been widely studied and fully applied in many computer vision fields, Szegedy et al. [13] still found interesting weaknesses in deep neural networks. Deep learning models are susceptible to attacks from adversarial samples. Such attacks will cause the neural network to completely change its prediction results, and this shortcoming will be used by attackers, which will have a great negative impact on the reliability and security of the ReID system. These adversarial samples have only added a slight perturbation, so that the human visual system cannot detect this perturbation. The far-reaching implications of such phenomena have attracted many researchers to conduct research in the field of adversarial attack and defense. To examine the robustness of ReID systems is rather important because the insecurity of ReID systems may cause severe losses. For example, criminals may use the adversarial perturbations to cheat the monitoring systems. As far as we know, the defense research against attacks based on deep neural network ReID systems has just begun. [10, 32] are the only two current research results on adversarial attacks based on deep learning in ReID field, which shows that ReID also faces the above-mentioned problems. Therefore, it is an urgent problem to develop an effective ReID adversarial defense method. Our work not only provides a simple and effective adversarial defense method for ReID, but also improves the performance of the model. In terms of model accuracy improvement, Multi-Modal Data Augmentation enables the model to fully learn the spatial structure features, which not only strengthens the learning of the underlying features, but also further alleviates the over-fitting risk of ReID model because of the color variation caused, so as to further improve the accuracy of the model. In terms of adversarial defense, images of different modalities retain the homogeneous structure information of visible images, and the consistency of structural information ensures that the model can be trained and converged normally, the fusion of the different modal information can bring rich structural variation to the input data, which will make attack method difficult to align the adversarial space of the dataset. This is equivalent to the original adversarial space being disturbed, which means to produce defense effects. Based on the above analysis, we propose the two methods to effectively improve model accuracy and defensive ability: (1) multi-modal data augmentation, otherwise called Grayscale Patch Replacement, which is used to mine more robust features and enrich the input data variety, not only does it help to improve the performance of the model, but also brings robust defense capability to the model, so this method fully play dual potential of multi-modal information; (2) Multi-Modal Defense, which integrates three homogeneous modal images of visible, grayscale and sketch, and enriches the changes of input data and strengthens the defense ability of the model.

The proposed method has the following advantages:

(1) It is a lightweight approach which does not require any additional parameter learning or memory consumption. It can be combined with various CNN models without changing the learning strategy.

(2) It is a complementary approach to existing data augmentation and defense methods. When these methods are used altogether, it will further improve ReID accuracy and adversarial defense ability.

Our main contributions are summarized as follows:

(1)An effective method is proposed in this paper, which will make full use of the structural information of homogeneous modal images and increase the number and diversity of training samples. Hence, it will enhance the model's robustness to color variation. We have conducted a large number of experiments on three ReID datasets, which show the effectiveness of the proposed method.

(2) Using image scaling and the homogeneity of multi-modal information to perturb adversarial space, the combination method of internal and external defenses provides an effective growth direction for future research on improving the accuracy and defense ability of ReID models, and verifies the effectiveness and versatility of the method in a variety of attack methods and fields.

The content of the remainder of the paper is as follows: Section 2 presents the related work, Section 3 presents the related algorithms, Section 4 presents the experimental results and analysis, and Section 5 presents a conclusion.

## 2    Related Work

It's well known that data augmentation such as Random Cropping and Random Flipping plays an important role in the field of classification, detection and ReID. All of which increase the diversity of training data, and improve the generalization ability of the model to some extent. The Random Erasing proposed by Zhong et al. [3] simulates the occlusion problem that is frequently encountered in reality, which randomly erases a part of the image with a certain probability in the training samples to increase the diversity of the training samples. To some extent, it resolves the problems of inadequate generalization ability when the recognition task faces the occlusion problem, so it has become an effective training trick in the field of ReID. Zheng et al. used generative adversarial networks to change the clothes of a pedestrian with the clothes of other pedestrians, this method generates more diversified data and improves the generalization ability of the model [29]. The k-reciprocal encoding is used to reorder the results of the query to improve mAP and Rank-1 [34], this trick is known as re-Rank. The above methods partly solve the problem from different views, and effectively help the model to improve the accuracy. We propose a multi-modal data augmentation called Grayscale Patch Replacement, which uses spatial structure information and color information of visible and grayscale modal images to fit each other in the shared space, this method helps the model learn more effective features and enhance robustness. The following cross-domain experiments and adversarial defense experiments have fully shown that this method effectively improves the robustness of the model.

To achieve the goal of defense, Guo et al. [14] proposed to adopt more diversified non-differentiable image transformation operations such as depth reduction, JPEG compression, total variance minimization and image quilting to increase the difficulty of network gradient prediction to achieve the purpose of defense. It is noted that most of the training images are in JPG format, Dziugaite [1] uses JPG image compression method to reduce the impact of adversarial disturbance experimental results show that this method is effective for some adversarial attack algorithms. PixelDefense defense [27] uses the PixelCNN generation model to change all pixels in each channel and convert the adversarial sample to the normal sample then put in the original model prediction. VectorDefense[28] converts the bitmap into a vector image space and returns it before classification to avoid being deceived by the adversarial structure. These methods are not simple enough and have not been verified in more fields.

The performance of a ReID algorithm in an adversarial attack can evaluate the robustness of an algorithm. MS-SSIM [10] is an attack method specially proposed for ReID in CVPR2020. It has extremely strong transferable attack capability, and its transferable attack can make the model accuracy drop sharply. At the same time, it has also been demonstrated to be more aggressive than the traditional attack methods such as DeepFool [18], NewtonFool [15], CW [31]. As far as we know, this attack method currently has no better defense method. Metric Attack [32] is another ReID adversarial attack method, which uses distance measurement to achieve an attack on the ReID model. The Multi-Modal Defense we proposed not only has a good defensive effect on the above two ReID attack methods, but is also effective against other traditional attack methods. It is a universal defense method.

## 3 Proposed Method

### 3.1 Introduction to Multi-Modal Data Augmentation

Inspired by Zheng [29], this paper proposes an effective ReID multi-modal data augmentation method, which consists of Local Grayscale Patch Replacement(LGPR) and Global Grayscale Patch Replacement(GGPR). As illustrated in Figure 1 and Figure 2, the input image is randomly replaced with grayscale patch by of homogeneous grayscale ima- ge to make the model more fully mining the underlying features related to the spatial structure, which find a better balance between the color information and the structure information to enhance the generalization ability of the model.

We conduct experiments in subsections 3.1.1 and 3.1.2 using reID_baseline [30] on the Market-1501[9] dataset.

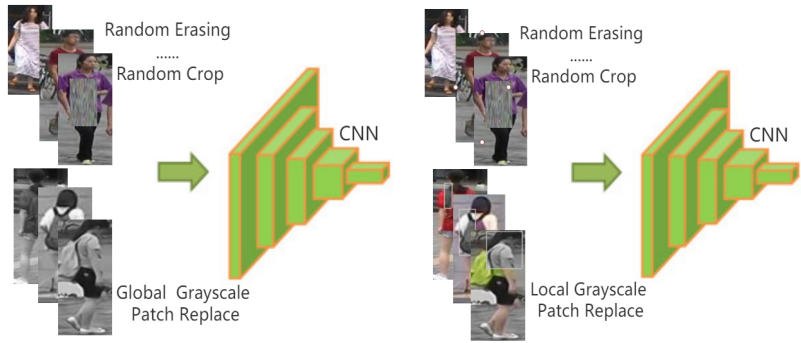

Figure 1: Application of GGPR(left) and LGPR(right) in pipelines of the baseline.

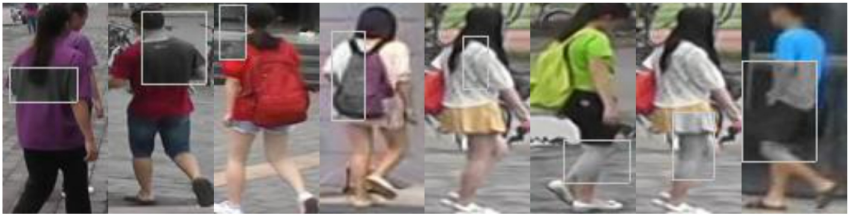

Figure 2: Schematic diagram of Random Grayscale Patch Replacement.

### 3.1.1 Introduction to Multi-Modal Data Augmentation

In the process of model training, we conduct LGPR randomly transformation on the entire batch of training images with a certain probability. For an image I in a mini-batch, denote the probability of it undergoing LGPR be $p_r$, and the probability of it being kept unchanged be $1-p_r$. In this process, it randomly selects a rectangular region in the image and replaces it with the pixels of the same rectangular region in the corresponding grayscale image. Thus, training images which include regions with different levels of grayscale are generated. Among them, $s_l$ and $s_h$ are the minimum and maximum values of the ratio of the image to the randomly generated rectangle area, and the $S_e$ of the rectangle area limited between the minimum and maximum ratio is obtained by $S_e \leftarrow$ Rand $(s_l, s_h) \times S$, $r_e$ is a coefficient used to determine the shape of the rectangle. It is limited to the interval $(r_1, r_2)$. $x_e$ and $y_e$ are randomly generated by coordinates of the upper left corner of the rectangle. If the coordinates of the rectangle exceed the scope of the image, the area and position coordinates of the rectangle are re-determined. When a rectangle that meets the above requirements is found, the pixel values of the selected region are replaced by the corresponding rectangular region on the grayscale image converted from RGB image. The procedure of LGPR is shown in Algorithm.1.

**Hyper-Parameter Setting**. During CNN training, there is a hyper-parameter that needs to be estimated here, that is, the random grayscale probability $p_l$. We take the hyperparameter $p_l$ as 0.01, 0.03, 0.05, 0.07, 0.1, 0.2, 0.3,..., 1 for the LGPR experiments. Then we take the value of each parameter for three independent repetitions of the experiments. Finally, we calculate the average of the final result. The results of different $p_r$ are shown in Figure 3.

Obviously, when $p_l$=0.4 or $p_l$=0.7, the model achieves better performance. And the best performance is achieved when pl=0.4. If we do not specify, the hyper-parameter is set pl=0.4 in the later experiments.

**Algorithm1: Local Grayscale Patch Replace Procedure**

**Input:** Input image I;
 Image size W and H;
 Area of image S;
 Erasing probability p;
 Erasing area ratio range $s_l$ and $s_h$ ;
 Erasing aspect ratio range $r_1$ and $r_2$ .
**Output:** Grayscale erased image $I^*$ .
1 **Initialization:** $p_1 \leftarrow$ Rand (0, 1).
2 **if** $p_1 \geq p$ **then**
3 | $I^* \leftarrow I$;
4 | **return** $I^*$ .
5 **else**
6 | **while** True **do**
7 | | $S_e \leftarrow$ Rand ($s_l$ ,$s_h$ )×S;
8 | | $r_e \leftarrow$ Rand ($r_1$ ,$r_2$ );
9 | | $H_e \leftarrow$ Sqrt($S_e$×$r_e$), $W_e \leftarrow$ Sqrt($S_e$ / $r_e$);
10 | | $x_e \leftarrow$ Rand (0,W), $y_e \leftarrow$ Rand (0,H);
11 | | **if** $x_e + W_e \leq W$ and $y_e + H_e \leq H$ **then**
12 | | | $I_e \leftarrow (x_e ,y_e ,x_e + W_e ,y_e + H_e )$;
13 | | | $I(I_e ) \leftarrow$ Grayscale($I_e$);
14 | | | $I^* \leftarrow I$;
15 | | | **return** $I^*$
16 | | **end**
17 **end**
18

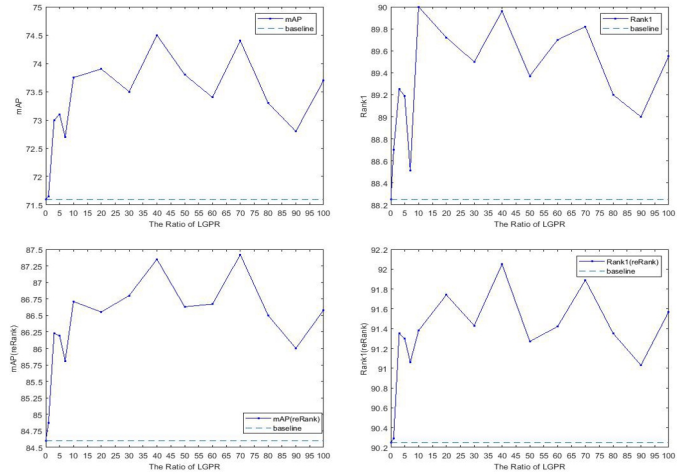

Figure 3: Accuracy under different hyper-parameters on Market-1501 with using baseline[30].

**3.1.2. Global Grayscale Patch Replace(GGPR)**

In the process of model training, we conduct GGPR randomly transformation on the entire batch of training images with a certain probability. When the probability value of GGPR is set to p=0.05, it means that 5% of the input samples are converted into grayscale images to participate in the model training during the training process. This method can be implemented by the RandomGrayscale function in pytorch.

**Hyper-Parameter Setting**. During CNN training, there is a hyperparameter that needs to be estimated here, that is, the random grayscale probability $p_g$. We take the hyper-parameter $p_g$ as 0.01, 0.03, 0.05, 0.07, 0.1, 0.2, 0.3, ... , 1 for the GGPR experiments. Then we take the value of each parameter for three independent repetitions of the experiments. Finally, we calculate the average of the final result.The results of different $p_g$ are shown in Figure 4.

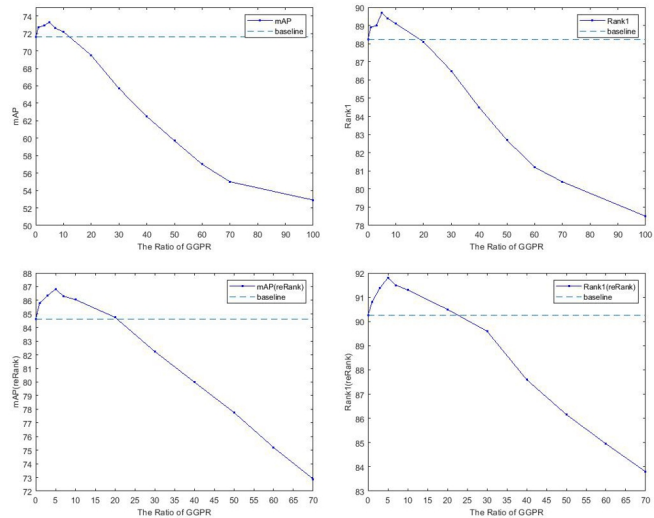

Figure 4: Accuracy under different hyper-parameters on Market-1501 using baseline[10].

We can see that when $p_g$=0.05, the performance of the model reaches the maximum value in Rank-1 and mAP in Figure.4. If we do not specify, the hyper-parameter is set $p_g$=0.05 in the later experiments.

**Evaluation of GGPR and LGPR**. Compared with the best results of GGPR on baseline [10], the accuracy of LG- PR is improved by 0.5% and 1.4% on Rank-1 and mAP, re- spectively. Under the same conditions using re-Rank, the accuracy of Rank-1 and mAP is improved by 0% and 0.4%, respectively. Therefore, the advantages of LGPR are more obvious when re-Rank is not used. However, Figure 3 also shows that the performance improvement brought by LGPR is not stable enough because of the obvious fluctuation in LGPR, while the performance improvement brought by GGPR is very stable. Therefore, we improve the stability of the method by combining GGPR with LGPR.

**Evaluation by Combining GGPR with LGPR**. First, we fix the hyper-parameter value of GGPR to $p_g$=0.05, then keep the control variable unchanged to further determine the hyper-parameter of LGPR. Finally, we take the hyper- parameter $p_l$ of LGPR to be 0.1, 0.2, ... , 0.7 to conduct combination experiments of GGPR and LGPR, and cond- uct three independent repeated experiments for each parameter $p_l$ to get the average value. The result is shown in Figure 5. It can be seen from Figure that the performance improvement brought by the combination of GGPR and LGPR is more stable and with less fluctuation, and the comprehensive performance of the model is the best when the hyper-parameter value of LGPR is $p_l$=0.4.

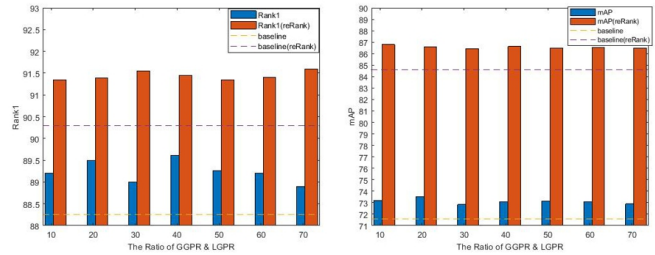

Figure 5. Model performance of combining GGPR with LGPR.

### 3.2 Multi-Modal Defense

A grayscale image can be obtained when a visible image extracts color information, and a sketch image can be obtained when a grayscale image extracts details information. In other words, a grayscale image is a visible image that loses some color information, and a sketch image is a grayscale image that loses some details information. Thus, visible images, grayscale images, and sketch images are homogeneous.

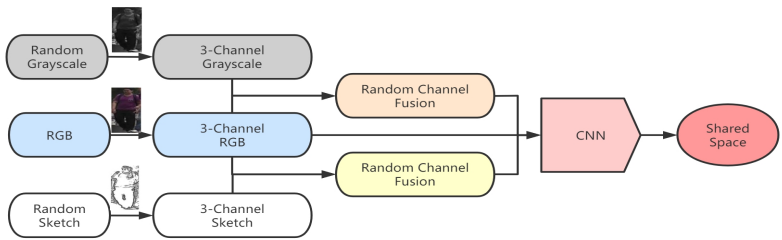

Figure 6: The pipeline of Multi-Modal Defense of proposed method.

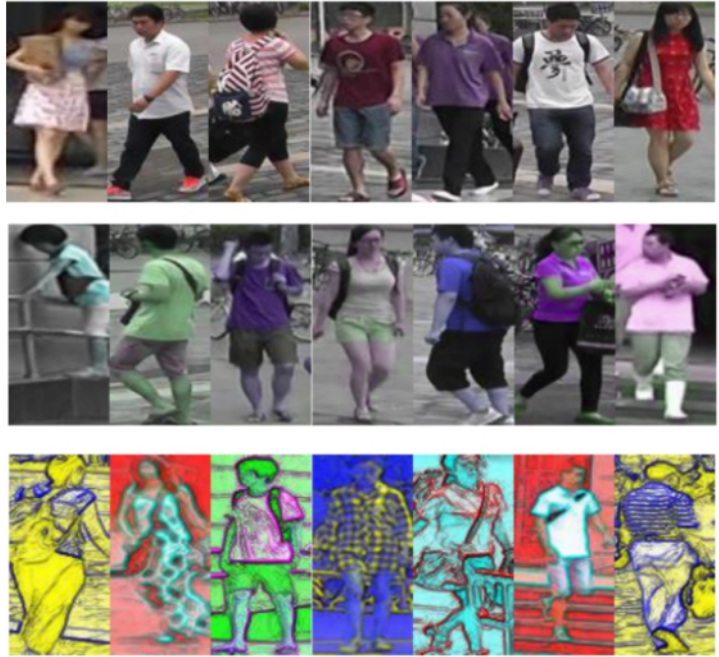

Figure 7: The first row of picture is an RGB image, the second row of picture is a partial display of the random fusion of RGB image and grayscale image channels, and the third row of picture is a par- tial display of the random fusion of RGB image and sketch image channels. We can see that our method has rich changes by fusing RGB image with grayscale and sketch image.

Our Multi-Modal Defense solution is to train the model using visible modal and homogeneous grayscale modal and sketch modal images. As shown in Figure 6, we randomly convert a batch of RGB images into grayscale or sketch images with a certain probability in pre-processing stage, and then merge the channels of the grayscale image and the sketch image with the channels of the RGB image. The advantages of using homogeneous modal images can be attributed to two aspects: (1) The grayscale and sketch images retain the homogeneous structural information of the visible image. It is the key to ensure the convergence of the model. The sketch and grayscale images are different modalities of homogeneous RGB images, so using them for mixed training will not bring negative impact on model performance. (2) The proposed method can simply and efficiently bring rich changes to the input data, and still maintain good performance under a variety of attacks. The procedure is shown in Algorithm.2.

---
**Algorithm2 : Multi-Modal Defense Procedure**
---

**Input :**  Input batch RGB images I;
  Gray transformation probability $p_1$;
  Sketch transformation probability $p_2$;
**Output1:** batch of grayscale images $I_1$.
**Output2:** batch of RGB-Gray fusion images $I_2$.
**Output3:** batch of RGB-Sketch fusion images $I_3$.
**Initialization:** $p_1 \leftarrow 0.1$; $p_2 \leftarrow 0.05$; $p_3 \leftarrow 0.05$.
**1 if is_train:**
**2**   if $p_1+p_2+p_3 \geq p$ then
**3**     If $p_2+p_3 \geq p \geq p_3$
**4**       $I_2 \leftarrow I$;
**5**     If $p_2 \geq p$
**6**       $I_3 \leftarrow I$;
**7**   else
**8**     $I_1 \leftarrow I$;
**9 return I.**

---

We transform 15% and 5% RGB images into grayscale and sketch images, respectively, then we take two-thirds grayscale images as GGPR to directly input CNN. Meanwhile, we fuse the remaining grayscale and sketch image with RGB image. In the process of image fusion, we use the grayscale or sketch image to overlay 1 or 2 channels from RGB image channels randomly selected. The fusion effect is shown in Figure 7.

## 4    Experiments Results and Analysis

Datasets. We conducted comparison experiments on MTMC17[24], DukeMTMC[45], and Market-1501[9]. The MSMT17 dataset, created in winter, was presented in 2018 as a new, larger dataset closer to real-life scenes, containing a total of 4,101 individuals and covering multiple scenes and time periods. The DukeMTMC is a large-scale Multi-Target, Multi-Camera (MTMC) tracking dataset, a HD video dataset recorded by 8 synchronous cameras, with more than 7,000 single camera tracks and more than 2,700 individual pedestrians. The Market-1501 dataset was collected in the summer of 2015. It includes 1,501 pedestrians captured by six cameras (five HD cameras and one low-definition camera). These three datasets are currently the largest datasets of ReID, and they are also the most representative because they collectively contain multi-season, multi-time, HD, and low-definition cameras with rich scenes and backgrounds as well as complex lighting variations. We evaluated these three datasets using Rank-k precision and mean Average Precision(mAP). Rank-1 denotes the average accuracy of the first returned result corresponding to each query image; mAP denotes the mean of average accuracy, the query results are sorted according to the similarity, the closer the correct result is to the top of the list, the higher the score.

## 4.1 Comparison of Grayscale Patch Replacement and State-of-the-Art

In this section we will compare the performance of our approach with state-of-the-art methods on three baselines. The baselines are the ReID_baseline[30], the strong baseline [11] and FastReID[12]. Since the model requires more training epochs to fit than the original, we add 1 to 2 times training epochs to the training process. A comparison of the performance of our method with the state-of-the-art methods in three datasets is shown in Table 1 to Table 3.

To our knowledge, applying the proposed approach to fastReID, we have achieved the highest retrieval accuracy currently available on the MTMC17 dataset.

On the one hand, our method achieves better ReID performance because of exploiting the grayscale transformation, which increases the number and diversity of training samples. On the other hand, exploiting the structural information retained by the grayscale image, the colors of the RGB images and the spatial structural informion of the grayscale images are fitted to each other in the model training, increasing the model's robustness to color changes.

| Methods | MTMC17 | |
|---|---|---|
| | Rank-1 | mAP |
| IANet[19] (CVPR'19) | 75.5 | 46.8 |
| DGNet[29](CVPR'19) | 77.2 | 52.3 |
| OSNet[21](ICCV'19) | 78.7 | 52.9 |
| BATNet[25](ICCV'19) | 79.5 | 56.8 |
| RGA-SC[26](CVPR'20) | 80.3 | 57.5 |
| SCSN[6](CVPR'20) | 83.8 | 58.5 |
| AdaptiveReID[35](arXiv'20) | 81.7 | 62.2 |
| fastReID[12] | 85.1 | 63.3 |
| [12] + GGPR(ours) | **86.2(+1.1)** | **65.3(+2)** |
| [12] + GGPR&LGPR(ours) | **86.2(+1.1)** | **65.9(+2.6)** |

**Table 1**: Performance comparison on MTMC17 datasets. Since using reRank on the MTMC17 dataset requires a lot of hardware cost, we did not make a comparison in this regard.

| Methods | Market1501 | |
|---|---|---|
| | Rank-1 | mAP |
| PCB [36] (ECCV'18) | 92.3 | 77.4 |
| AANet [40] (CVPR'19) | 93.9 | 83.4 |
| IANet [19] (CVPR'19) | 94.4 | 83.1 |
| Auto-ReID [44] (ICCV'19) | 94.5 | 85.1 |
| DGNet[29](CVPR'19) | 94.8 | 86.0 |
| Pyramid [42] (CVPR'19) | 95.7 | 88.2 |
| ABDNet [43] (ICCV'19) | 95.6 | 88.3 |
| SONA [38] (ICCV'19) | 95.7 | 88.7 |
| SCAL [39] (ICCV'19) | 95.8 | 89.3 |
| CAR [37] (ICCV'19) | 96.1 | 84.7 |
| Circle Loss [41] (CVPR'20) | 96.1 | 87.4 |
| ReID_baseline [30] | 88.8 | 71.6 |
| [30] + reRank | 90.5 | 85.2 |
| [30] + GGPR(ours) | **89.5(+0.7)** | **73.5(+1.9)** |
| [30] + GGPR+ reRank(ours) | **92.0(+1.5)** | **86.9(+1.7)** |
| [30] + LGPR(ours) | **90.0(+1.2)** | **74.9(+3.3)** |
| [30] + LGPR + reRank(ours) | **92.0(+1.5)** | **87.4(+2.2)** |
| strong baseline [11] | 94.5 | 85.9 |
| [11] + reRank | 95.4 | 94.2 |
| [11] + GGPR(ours) | **94.6(+0.1)** | 85.7 |
| [11] + GGPR+ reRank(ours) | **96.2(+0.8)** | **94.7(+0.5)** |
| [11] + LGPR(ours) | **95.1(+0.6)** | **87.2(+1.3)** |
| [11] + LGPR + reRank(ours) | **95.9(+0.5)** | **94.4(+0.2)** |
| fastReID [12] | 96.3 | 90.3 |
| [12] + reRank | 96.8 | 95.3 |
| [12] + GGPR(ours) | **96.5(+0.2)** | **91.2(+0.9)** |
| [12] + GGPR+ reRank(ours) | **96.9(+0.1)** | **95.6(+0.3)** |

**Table 2**: Performance comparison on Market1501 datasets.

| Methods | DukeMTMC | |
|---|---|---|
| | **Rank-1** | **mAP** |
| PCB [36] (ECCV'18) | 81.8 | 66.1 |
| AANet [40] (CVPR'19) | 87.7 | 74.3 |
| IANet [19] (CVPR'19) | 87.1 | 73.4 |
| Auto-ReID [44] (ICCV'19) | — | — |
| DGNet[29](CVPR'19) | 86.6 | 74.8 |
| Pyramid [42] (CVPR'19) | 89.0 | 79.0 |
| ABDNet [43] (ICCV'19) | 89.0 | 78.6 |
| SONA [38] (ICCV'19) | 89.3 | 78.1 |
| SCAL [39] (ICCV'19) | 89.0 | 79.6 |
| CAR [37] (ICCV'19) | 86.3 | 73.1 |
| Circle Loss [41] (CVPR'20) | 89.0 | 79.6 |
| strong baseline [11] | 86.4 | 76.4 |
| [11] + reRank | 90.3 | 89.1 |
| [11] + GGPR(ours) | **87.8(+1.4)** | **77.3(+0.9)** |
| [11] + GGPR+ reRank(ours) | **90.9(+0.6)** | **89.2(+0.1)** |
| [11] + LGPR(ours) | **87.3(+0.9)** | **77.3(+0.9)** |
| [11] + LGPR + reRank(ours) | **91(+0.7)** | **89.4(+0.3)** |
| fastReID [12] | 92.4 | 83.2 |
| [12] + reRank | 94.4 | 92.2 |
| [12] + LGPR(ours) | **92.8(+0.4)** | **84.2(+1)** |
| [12] + LGPR + reRank(ours) | 94.3 | **92.7(+0.5)** |

**Table 3**: Performance comparison on DukeMTMC datasets.

## 4.2 Cross-domain Person Re-identification

Cross-domain person re-identification aims at adapting the model trained on a labeled source domain dataset to another target domain dataset without any annotation. The performance of a method in cross-domain experiments can test whether the method really improves the robustness of the model. In order to further explore the effectiveness of the proposed method in cross-domain experiments, we use GGPR to conduct the following cross-domain experiments on strong baseline [11]. The experiments are shown in Table 4. Experimental results show that Random Erasing can also significantly improve the performance of the ReID model, but it will cause a significant drop in cross-domain performance. The proposed method can not only signific- antly improve the cross-domain performance of the ReID model, but also be more robust because of learning more discriminative features.

In Table 4, **+REA** means that the trick of Random Erasing is used in model training, **-REA** means turning it off.

| Methods | M→D | | D→M | |
|---|---|---|---|---|
| | Rank-1 | mAP | Rank-1 | mAP |
| [11]+REA+reRank | 33.6 | 24.3 | 51.6 | 32.3 |
| [11]+REA+GGPR+reRank(ours) | **37.8** | **27.8** | **55.4** | **35.7** |
| [11]-REA+reRank | 45.5 | 37.0 | 58.2 | 37.8 |
| [11]-REA+GGPR+reRank(ours) | **48.2** | **37.9** | **65.0** | **43.7** |

**Table 4**: The performance of different models is evaluated on cross-domain dataset. M→D means that we train the model on Market1501 and evaluate it on DukeMTMC.

## 4.3 Adversarial Defense

We use the adversarial samples generated by the MS-SSIM attack on AlignedReID [2] to conduct migration attack defense experiments. Meanwhile, we use the adversarial samples generated by the Metric Attack attack on HACNN [33] to conduct another migration attack defense experiment. The targets of the two migration attacks are all strong baseline [11]. The following experiment adopts the result of using reRank by default.

### 4.3.1. Use GGPR to Improve Adversarial Defense

We use GGPR on the Market-1501 dataset to conduct the following adversarial defense experiments on strong base- line [11].

**Experimental details**. We use scaling to conduct the ad- versarial defense experiments. The image size of the Market-1501 dataset is 128*64, and the strong baseline [11] will rescale the image size to 384*128 during the training proc- ess. We first rescale the query image to 100*50, and then rescale it to 384*128 in the inference stage, this operation will destroy the noise structure of the adversarial sample from the exterior of the model without significantly harm- ing the model performance. Finally, using above proposed method to defend the model interior, the images rescaled are input to the model. It is notable that we only use image scaling preprocessing in the inference stage. The experim- ental results show that this combination of internal and ext- ernal defenses works well.

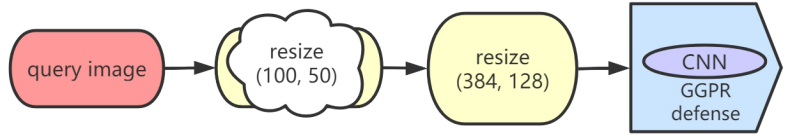

Figure 8: Schematic diagram of internal and external combined defense.

The MS-SSIM attack and defensive experimental results are shown in Table 5:

| Methods | Market1501 | | | |
| --- | --- | --- | --- | --- |
| | clean | | MS-SSIM (epoch=30) | |
| | Rank-1 | mAP | Rank-1 | mAP |
| reID-strong baseline[11] | 95.4 | 94.2 | 0.2 | 0.8 |
| **[11]+GGPR(Ours)** | **96.2** | **94.7** | **92.0** | **89.9** |

**Table 5**. The MS-SSIM attack and defensive experiment.

**Defensive Performance Analysis**. After learning multi-modal homogeneous data, we think that the model will ma- ke attack method difficult to align this change in the adversarial space of the dataset, so that the model trained by this method produces defense effects. To further experiments, we use more multi-modal to conduct defense experiments.

### 4.3.2. Multi-Modal Defense Experiment

In the MS-SSIM defense experiment, we rescale an image to 110*50 in the pre-processing stage. Comparison of the defense effects of GGPR and Multi-Modal Defense methods under epoch=30 and epoch=40 of iterative attack conditions are shown in Table.6. We can see from the Table that only using image scaling and GGPR technology can achieve a better defense effect, and the combination of using scaling and Multi-Modal Defense can further improve the defense effect. It is remarkable that the adversarial noise of the adversarial samples generated by MS-SSIM is already visible to the human visual system under the condition of epoch=40, and the proposed method can still defend well the attack.

| Methods | Market1501 | | | |
|---|---|---|---|---|
| | clean | | MS-SSIM (epoch=30) | |
| | Rank-1 | mAP | Rank-1 | mAP |
| AlignedReid[2] | 94.0 | 91.2 | 0.1 | 0.3 |
| reID-strong baseline[11] | 95.4 | 94.2 | 0.2 | 0.8 |
| [11] + Ours[1] | **96.2(+0.8)** | **94.7(+0.5)** | **92.0(+92)** | **89.9(+89.1)** |
| [11] + Ours[2] | **95.8(+0.4)** | 94.0 | **93.3(+93.1)** | **91.2(+90.4)** |

| Methods | Market1501 | | | |
|---|---|---|---|---|
| | clean | | MS-SSIM (epoch=40) | |
| | Rank-1 | mAP | Rank-1 | mAP |
| AlignedReid[2] | 94.0 | 91.2 | 0.0 | 0.2 |
| reID-strong baseline[11] | 95.4 | 94.2 | 0.2 | 0.6 |
| [11] + resize[100,50] | 95.1 | 93.9 | **63.5(+63.3)** | **63.0(+62.4)** |
| [11] + Ours[1] | **96.2(+0.8)** | **94.7(+0.5)** | **77.4(+77.2)** | **75.8(+75.2)** |
| [11] + Ours[2] | **95.8(+0.4)** | 94.0 | **90.9(+90.7)** | **89.2(+88.6)** |

**Table 6**: Comparison of the defense effects of Ours[1]: GGPR and Ours[2]: Multi-Modal Defense.

Since Adversarial samples are generated from the query set by MS-SSIM, and adversarial samples are generated from the gallary set by Metric Attack. For fair comparison, we sample another query set from the gallary adversarial samples of the Metric Attack attack, and the size of the adversarial samples are all 256*128, and the adversarial defense experiment is conducted on the strong baseline [11]. In the Metric Attack defense experiment, we rescale the training samples to 110*50 in preprocessing stage. The comparison of the defense effects of GGPR and Multi-Modal Defense methods against Metric Attack attacks is shown in Table 7:

| Methods | Market-1501[9] | | | |
|---|---|---|---|---|
| | clean | | Metric Attack | |
| | Rank-1 | mAP | Rank-1 | mAP |
| [attack gallary] HACNN[33](no reRank) | 91.3 | 77.5 | 0.0 | 0.0 |
| [attack query] reID-strong baseline[11] | 93.5 | 92.9 | 56.7 | 62.6 |
| [11]+resize[110,50] | 92.8 | 92.3 | **68.4(+11.7)** | **72.4(9.8)** |
| [11]+Ours[1] | **94.1(+0.6)** | **93.1(+0.2)** | **77.8(+21.1)** | **79.3(+16.7)** |
| [11]+Ours[2] | **94.1(+0.6)** | 92.8 | **87.2(+30.5)** | **86.1(+23.5)** |

**Table 7**: Comparison of the defense effects of GGPR and Multi-Modal Defense in Metric Attack.

## 5    Conclusion

In this paper, a simple and effective ReID data augmentation method with adversarial defense effects is proposed. Neither does the method require large scale training like GAN, nor introduces any noise. This method increases the number and diversity of training samples through homogeneous grayscale replacement, and avoids over-fitting the color variations introduced during the training process. The effectiveness of the proposed method is verified by experiments on multiple datasets and baselines, and exceeds the performance of the currently optimal algorithm. In addition, the proposed method uses image scaling to break the adversarial structure from the exterior of adversarial image, and then brings a variety of input data through the fusion of homogeneous RGB images, grayscale images and sketch images, which makes attack method difficult to align this change in the adversarial space of the dataset. These operations will ultimately realize the combined defense of internal and external. The effectiveness of the proposed method is verified by testing on the only two attack methods against ReID. Finally, we hope that this work can promote the development of deep learning adversarial defenses and reduce the security issues of reID, and also inspire future researchers.